\documentclass[letterpaper, 10 pt, conference]{ieeeconf}  

\IEEEoverridecommandlockouts                              

\usepackage{graphics} 
\usepackage{epsfig} 
\usepackage{mathptmx} 
\usepackage{times} 
\usepackage{amsmath} 
\usepackage{amssymb}  
\usepackage{booktabs}
\usepackage{multirow} 
\usepackage[table]{xcolor} 
\usepackage{cuted} 
\usepackage{caption} 
\usepackage{makecell}
\title{\LARGE \bf
Order Matters: On Parameter-Efficient Image-to-Video \\  Probing for Recognizing Nearly Symmetric Actions
}

\author{Thinesh Thiyakesan Ponbagavathi$^{1,2}$ and Alina Roitberg$^{2}$
\thanks{$^{1}$Institute for Artificial Intelligence, University of Stuttgart, Germany.
$^{2}$Intelligent Assistive Systems Lab, University of Hildesheim, Germany.}
}

\newcommand{\alina}[1]{\textcolor{purple}{#1}}

\begin{document}

\maketitle

\thispagestyle{empty}
\pagestyle{empty}




\begin{abstract}
Fine-grained understanding of human actions is essential for safe and intuitive human--robot interaction. 
We study the challenge of recognizing  \emph{nearly symmetric actions}, such as 
\emph{picking up} vs. \emph{placing down} a tool or \emph{opening} vs. \emph{closing} 
a drawer. 
These actions are common in close human-robot collaboration, yet they are rare and largely overlooked 
in mainstream vision frameworks. 
Pretrained vision foundation models (VFMs) are often adapted using \emph{probing}, valued in robotics for its efficiency and low data needs, or \emph{parameter-efficient fine-tuning} (PEFT), which adds temporal modeling through adapters or prompts.
However, our analysis shows that probing is permutation-invariant and blind to frame order, while PEFT is prone to overfitting on smaller HRI datasets, and less practical in real-world robotics due to compute constraints.


To address this, we introduce \textsc{Step} (\textbf{S}elf-attentive \textbf{T}emporal 
\textbf{E}mbedding \textbf{P}robing), a lightweight extension to probing that models temporal order via frame-wise positional encodings, a global CLS token, and a simplified attention block. 
Compared to conventional probing, \textsc{Step} improves 
accuracy by 4--10\% on nearly symmetric actions and 6--15\%  overall across action recognition benchmarks in human-robot-interaction, industrial assembly, and driver assistance. Beyond probing, \textsc{Step} surpasses heavier PEFT methods and even outperforms fully fine-tuned models on all three benchmarks, establishing a new state-of-the-art. Code and models will be made publicly available: https://github.com/th-nesh/STEP.
\end{abstract}




\section{Introduction}
\label{sec:intro}

Vision foundation models (VFMs) \cite{bommasani2021opportunities} are reshaping robotic perception \cite{robot_1,robot_2,robot_3}, offering transferable visual representations that generalize across diverse tasks.  In human–robot interaction (HRI), this is crucial: robots must reliably recognize human activities and anticipate intentions from subtle hand-object interactions \cite{hri_2,hri_1}. For example, in assembly, a robot may need to decide whether to hand over a workpiece or place it aside - decisions that hinge on accurate action recognition. 

 \begin{figure}[t!]
    \centering
\includegraphics[width=1\columnwidth]{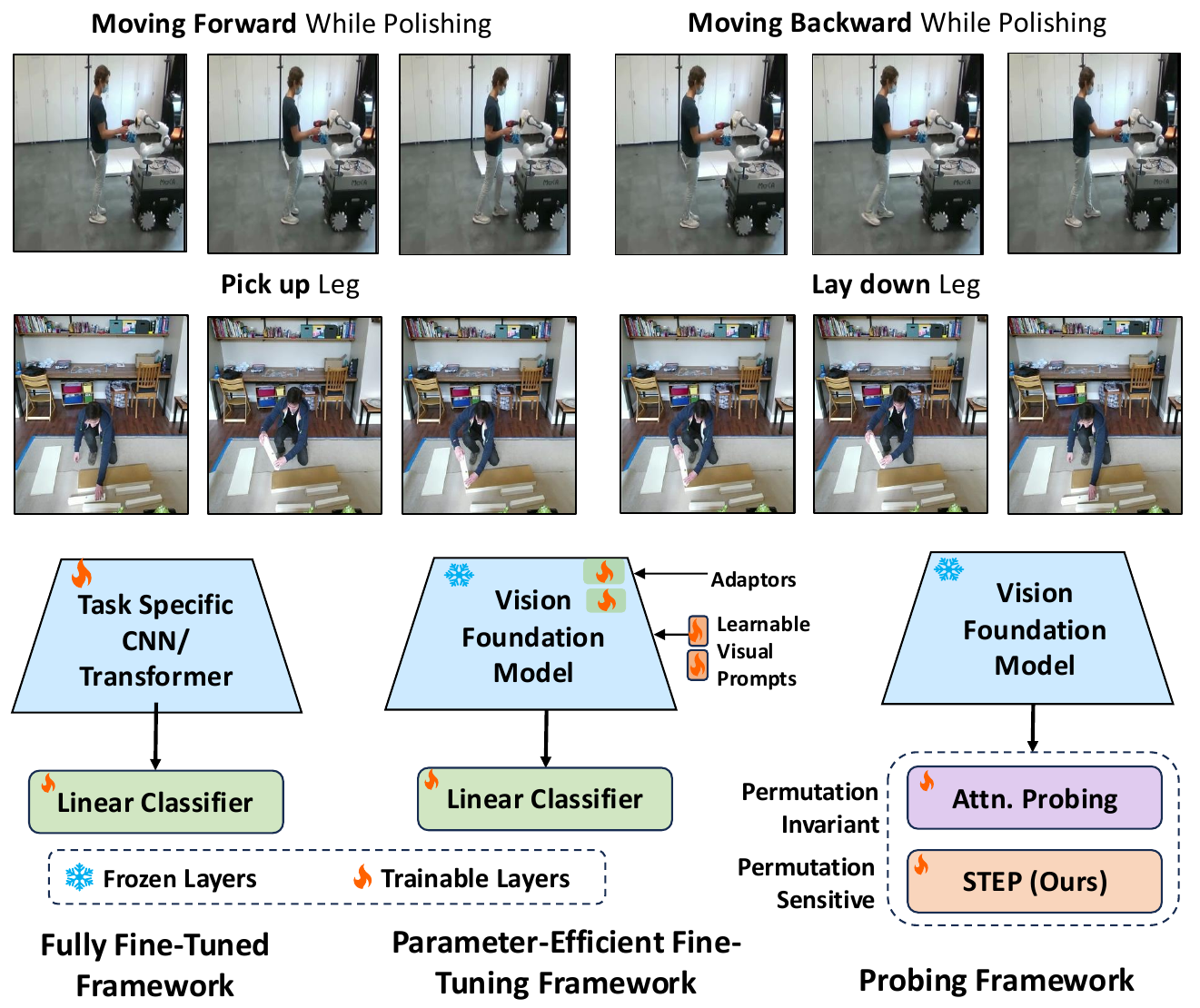}
\caption{\textbf{\textsc{Step}: Probing Vision Foundation Models for HRI}. \textsc{Step} makes probing (i.e., adding small task-specific heads on top of a frozen VFM backbone) sensitive to frame sequence, enabling accurate recognition of nearly symmetric actions (e.g., \textsl{pick up} vs. \textsl{lay down}) common in HRI scenarios. We focus on parameter-efficient VFM probing (right) and outperform parameter-heavier paradigms such as PEFT (middle) and task-specific models (left). 
}
\vspace{-2em}
\label{fig:teaser}

\end{figure}

Unlike large-scale activity benchmarks such as Kinetics \cite{kay2017kinetics400} or SSv2 \cite{ssv2}, HRI datasets \cite{hri_30,hri_1} are small, domain-specific, and often contain \textit{nearly symmetric actions} (eg, \textsl{putting down} vs. \textsl{picking up} or \textsl{opening} vs. \textsl{closing}). These actions look visually identical but differ in temporal order and their confusion can compromise safe and effective robot collaboration.
While prior studies explored order (e.g, arrow-of-time prediction \cite{arrow_1,arrow_2} and frame-order verification \cite{shuffle_1,shuffle_2}), they treated it mainly as an auxiliary signal. Consequently, evaluations that rely only on overall accuracy can obscure whether models truly capture temporal order. This limitation is particularly critical in HRI datasets, where 50–70\% of the categories are symmetric and reliable recognition depends on explicitly modeling sequence information.

However, adapting VFMs to capture temporal nuances in HRI is non-trivial. Two strategies dominate: probing and parameter-efficient fine-tuning (PEFT).
Probing \cite{coca,linear_prob_representationlearningcontrastivepredictive,set_transformer} freezes the backbone and trains a lightweight classifier on top. It is efficient and data-friendly but inherently permutation-invariant, ignoring frame order.
PEFT \cite{St_adaptor,VitaCLIP} inserts small trainable modules (adapters or prompts) into the frozen backbone to learn temporal dynamics. While more expressive, PEFT is heavier, prone to overfitting on small HRI datasets, and costly in multi-task settings where robots must solve several perception tasks simultaneously.   
Although CNN-based models \cite{hri_30,hri_cnn_1,hri_cnn_2} remain attractive in HRI for their efficiency, VFMs with probing offer a more scalable alternative by supporting multiple tasks in a single backbone pass. This motivates us to revisit probing vision foundation models for action recognition in HRI, while directly addressing its central weakness: weak temporal modeling.



Motivated by these challenges, we propose \textbf{Self-attentive Temporal Embedding Probing (\textsc{Step})} -- a lightweight extension of self-attention probing that introduces explicit temporal modeling at the probing stage through three components: (1) a learnable frame-wise positional encoding reinforcing the video’s sequential nature,  (2) a global CLS token shared across frames for temporal coherence, and (3) a simplified attention block with no skip connections, layer norm, or FF layers. Unlike PEFT, which modifies the backbone, \textsc{Step} keeps VFMs frozen and injects temporal order directly into the probing head, balancing efficiency and accuracy.
Evaluated on HRI-30 \cite{hri_30} (human-robot collaboration), IKEA-ASM \cite{IKEA_ASM} (furniture assembly), and Drive\&Act \cite{drive_and_act} (human-vehicle interaction), \textsc{Step} achieves state-of-the-art overall performance, surpassing both probing and PEFT baselines, with improvements of 6–15\% over probing baselines. Gains are especially pronounced on nearly symmetric actions, but \textsc{Step} also delivers the best overall accuracy across all datasets, confirming that explicit temporal order modeling is essential.

To summarize our contributions: 
  1) We explore the concept of \textit{nearly symmetric actions}: actions with visually similar frames but temporally opposite actions, and provide dedicated evaluations on three HRI benchmarks.
    2) We analyze the limitations of probing and PEFT in HRI scenarios comprising such nearly-symmetric actions (often fine-grained object manipulation tasks),  showing that probing is permutation-invariant to frame order, while PEFT overfits on smaller HRI datasets.
    3) We present \textsc{Step} -- a simple yet effective attention-based probing mechanism,  which integrates learnable frame-wise positional encodings, a frame-global CLS token, and a simplified attention block to better model temporal order. 
    4) Across benchmarks, \textsc{Step} achieves state-of-the-art accuracy with fewer parameters than probing and PEFT approaches, while also delivering substantial gains on nearly symmetric actions. 
    5) We further show that \textsc{Step} supports multi-task HRI in a single backbone pass, reducing computation up to 6× compared to PEFT.

\section{Related Work}
\label{sec:related_work}

\noindent\textbf{Action Recognition in HRI.}
Action recognition is central to HRI, enabling robots to understand human behavior and collaborate effectively. Earlier works relied on CNN-based pipelines,  typically two-stream models combining RGB and optical flow \cite{hri_cnn_2}, later extended with context-aware CNNs \cite{hri_cnn_context}, human–object parsing \cite{hri_cnn_1}, or CNN+LSTM backbones \cite{hri_cnn_3}. Digital twin \cite{hri_cnn4} frameworks and multimodal fusion \cite{hri_2} have also been explored, but all rely on CNNs as feature extractors.
While efficient on embedded hardware, these pipelines are inherently task-specific, and generalize poorly across domains. Vision Foundation Models \cite{CLIP, dinov2} (VFMs), by contrast, offer strong cross-task generalization but remain less popular in action recognition for HRI. Yet, despite their practical importance in HRI, \textit{nearly symmetric actions} remain largely unexplored in current pipelines. Thus, we explicitly study this regime and investigate how VFMs can be adapted to disambiguate such order-sensitive actions.

\noindent\textbf{Probing and PEFT for Adapting VFMs.} There are two popular ways for adapting image VFMs for action recognition: probing and parameter-efficient fine-tuning (PEFT). 
Linear probes \cite{linear_prob_representationlearningcontrastivepredictive,linear_prob_simclr} are simple but limited in expressiveness, while attentive probing with cross-attention \cite{coca,set_transformer} improves feature aggregation. In video tasks, probing typically averages or concatenates frame embeddings \cite{dinov2,video_glue,v-jepa}, or applies attentive probing \cite{coca,v-jepa,video_glue}, but these remain permutation-invariant \cite{set_transformer}, failing to capture temporal order. Parameter-Efficient Fine-Tuning (PEFT) adapts the backbone more deeply than probing. In the image domain, techniques such as Visual Prompt Tuning (VPT) \cite{vpt} introduce learnable prompts, while AdaptFormer \cite{adaptformer} inserts lightweight adapters into transformer layers to enable efficient adaptation. For video, ST-Adaptor \cite{St_adaptor} applies bottleneck layers for spatiotemporal transfer, 
Vita-CLIP \cite{VitaCLIP} leverages multimodal prompts, and M2-CLIP \cite{m2_clip} enhances temporal alignment with TED-Adapters. These methods achieve strong temporal modeling on large datasets, but risk overfitting in fine-grained HRI scenarios and scale poorly in multi-task robotics. This gap motivates our \textsc{Step} framework, leveraging probing efficiency but also modeling temporal dynamics for improved action recognition in HRI.

\noindent\textbf{Temporal Modeling for Image-to-Video Transfer.}
Temporal modeling in foundation models has evolved from frame-wise representations with learnable temporal encoders \cite{x-clip, ActionCLIP} to spatiotemporal fusion via cross-attention \cite{EVL}. 
Later works \cite{stan,bike,ila} refine temporal reasoning through auxiliary modules, bidirectional alignment, or temporal masks, while others embed temporal cues implicitly into input tokens via tube embeddings or spatiotemporal patches \cite{tong2022videomae,clip4clip,wang2024internvideo2scalingfoundationmodels}. TaskAdapter++ \cite{taskadaptor++} explored explicit order modeling, though mainly in text encoders. Beyond architecture design, prior work has also examined temporal order sensitivity itself: arrow-of-time prediction \cite{arrow_1,arrow_2}, frame-order verification \cite{shuffle_1,shuffle_2}, and Retro-Actions \cite{retro} showed that reversing or shuffling frames often causes little performance drop on large benchmarks, suggesting most models rely primarily on spatial cues. Overall, existing methods capture order only implicitly and are rarely tested on datasets with many symmetric actions. In contrast, we focus on explicit order modeling at the probing stage, making recognition sequence-sensitive while retaining efficiency.

\section{Self-attentive Temporal Embedding Probing for Recognizing Nearly Symmetric Actions}
\label{sec:methods}
Our goal is to advance temporal understanding in parameter-efficient image-to-video probing for human action recognition specifically for human-robot interaction (HRI) scenarios, where robots must distinguish subtle, order-dependent behaviors. 
A key challenge lies in \textit{nearly symmetric actions}: visually similar activities that differ only in temporal order (e.g., manipulation actions like \emph{picking}  vs. \emph{placing} an object). We first analyze the permutation invariance of self-attention blocks, then introduce \textsc{Step}, which strengthens temporal reasoning through frame-wise embeddings and a global CLS token. To assess its impact, we curate symmetric action splits across three datasets focusing on HRI and human-vehicle interaction and report results on both symmetric subsets and full benchmarks (Sec.~\ref{sec:setup}).

\noindent\textbf{Self Attention and Permutation Invariance.}
Self-attention \cite{vaswani2017attention} computes pairwise token relations via dot-product attention, producing outputs $z_i = \sum_j \alpha_{ij} v_j$ with weights $\alpha_{ij}$ normalized over all tokens. Because these weights are agnostic to input order, self-attention is inherently permutation-invariant~\cite{set_transformer}. While effective when appearance dominates, it fails in order-dependent cases, such as \textit{nearly symmetric actions}. Our results confirm this limitation: reversing frames yields almost no accuracy change (Table~\ref{tab:temporal_sensitivity}).

\subsection{Self-attentive Temporal Embedding Probing}
\label{sub_sec:STEP}
\begin{figure}[t!]
    \centering
\includegraphics[width=1\columnwidth]{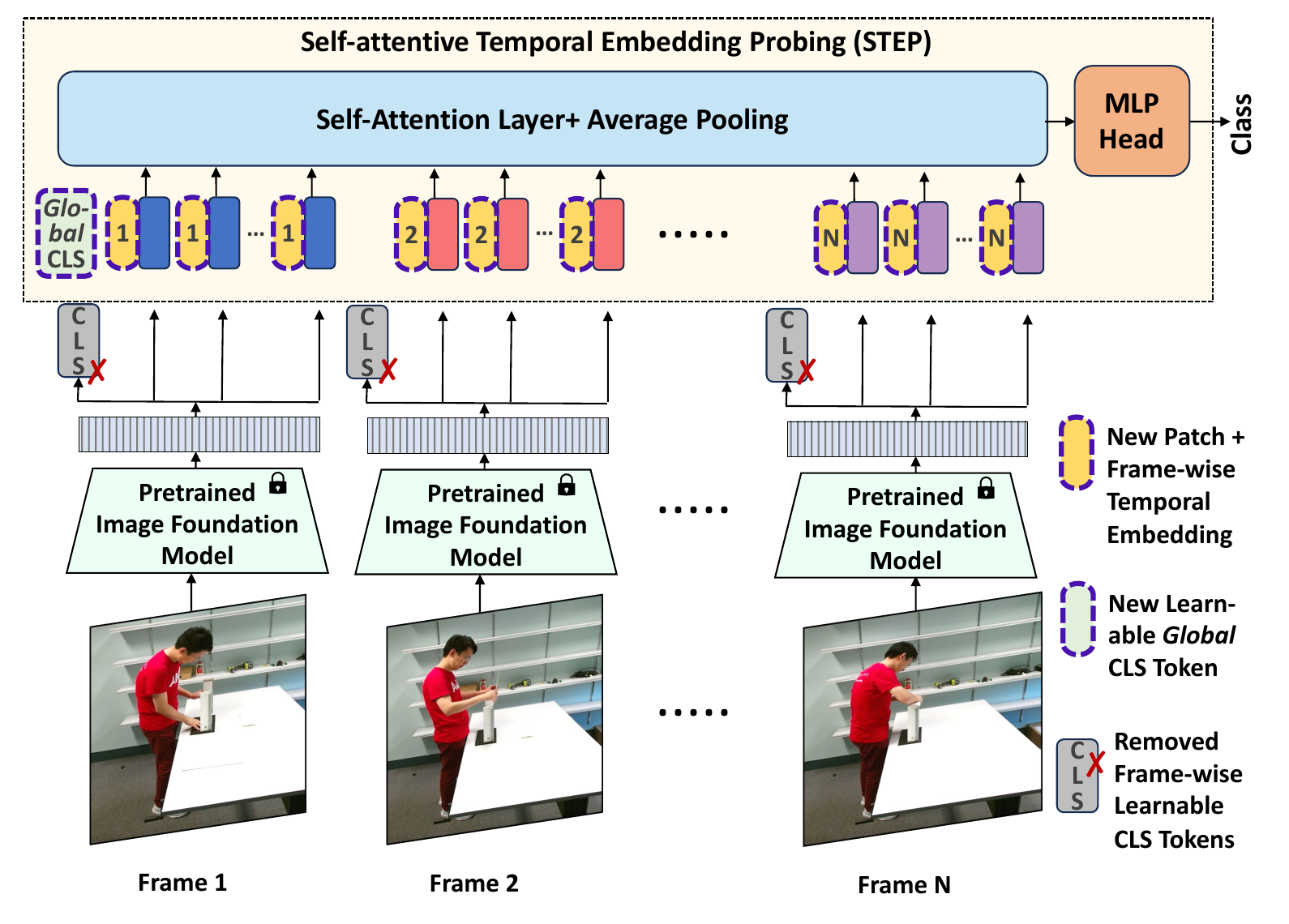}
\caption{\textbf{Overview of \textsc{Step}.} Each video frame is independently processed by a frozen image model. 
We replace the frame-specific CLS token with learned patch-wise temporal encodings,  while a newly added \textit{frame-global} CLS token encourages temporal consistency, followed by a self-attention probing mechanism that tracks temporal order. 
}

\label{fig:main}
\vspace{-1.5em}
\end{figure}

We now introduce \textbf{\textsc{Step}} - \textbf{S}elf-attentive \textbf{T}emporal \textbf{E}mbedding \textbf{P}robing (overview in Figure \ref{fig:main}), which incorporates several simple but effective modifications to attention-based probing tailored for sensitivity to subtle changes in temporal order, which often occur in close HRI tasks.
Consider a video sequence $x = \left\{ x_{1}, x_{2}, \ldots, x_{T} \right\}$  processed frame-by-frame with a pre-trained and frozen image foundation model $\theta_{\text{frame}}$, resulting in frame representations, $e_{i}$, $\theta_{\text{frame}}: \mathbf{x_{i}}\rightarrow e_{i}$. Each frame is spatially split into $n$ patches, forming a set of patch tokens $e_i^{\text{patch}} =\left\{ e_{i,1}, e_{i,2}, \ldots, e_{i,n} \right\}$ and an additional CLS token $e_i^{\text{CLS}}$ that provides a unified representation of the frame.


\begin{equation}
e_{i} =  \left\{ e_i^{\text{patch}}, e_i^{\text{CLS}} \right\}
\label{eq:embedding}
\end{equation}

The goal of our \textsc{Step} probing method, denoted as $f$, is then to link the frame-wise feature representations first into a sequence $e = \left\{ e_{1}, e_{2}, \ldots, e_{T} \right\}$ then mapping this sequence to $y$ -- the target action label:   $f: e\rightarrow y$.


Our method builds on the self-attention probing strategy for image-to-video transfer~\cite{clip4clip,set_transformer} with two main modifications: (1) a global CLS token that aggregates information across frames instead of the frame-wise CLS tokens, and (2) frame-wise temporal embeddings to encode temporal order. 
 These embeddings are then processed by a self-attention layer, average pooling, and a simplified classification head, as we consistently find that omitting certain common components of probing layers preserves action recognition accuracy while significantly reducing the parameter count.


\begin{table*}[ht!]
\centering
\resizebox{\linewidth}{!}{%
{
\begin{tabular}{l c cc|ccc|
@{\hspace{8pt}}|ccc|
@{\hspace{8pt}}|ccc}

\toprule

\textbf{Method} & \textbf{Backbone} & \textbf{GFLOPs} & \textbf{Train} 
& \multicolumn{6}{c|}{\textbf{Human-Robot Interaction}} 
& \multicolumn{3}{c}{\textbf{In-Vehicle Interaction}} \\

&&&\textbf{Params} 
& \multicolumn{3}{c}{\cellcolor{red!25}\textbf{HRI-30}} 
& \multicolumn{3}{c}{\cellcolor{blue!25}\textbf{IKEA-ASM}} 
& \multicolumn{3}{c}{\cellcolor{green!25}\textbf{Drive\&Act}} \\

&&&\textbf{(M)} 
& \textbf{Sym} & \textbf{N-Sym} & \textbf{Ovr.} 
& \textbf{Sym} & \textbf{N-Sym} & \textbf{Ovr.} 
& \textbf{Sym} & \textbf{N-Sym} & \textbf{Ovr.} \\

&&&& \textbf{Acc.} & \textbf{Acc.} & \textbf{Acc.} 
& \textbf{Acc.} & \textbf{Acc.} & \textbf{Acc.} 
& \textbf{Acc.} & \textbf{Acc.} & \textbf{Acc.} \\

\midrule
\addlinespace[2pt]
\multicolumn{13}{l}{\textbf{\textit{Comparison to published state-of-the-art (fully fine-tuned models)}}} \\
VideoSWINv2 \cite{liu2022video_swin}
& IN-21K   & 282& 88.1& - & - & - & - & - & 72.60 & - & - & - \\
\cellcolor{gray!20}SlowOnly \cite{hri_30} & \cellcolor{gray!20}K400 & \cellcolor{gray!20}75 &  \cellcolor{gray!20}32                    
& \cellcolor{gray!20}-&\cellcolor{gray!20} -& \cellcolor{gray!20}\underline{86.55} 
& \cellcolor{gray!20}-&\cellcolor{gray!20}-& \cellcolor{gray!20}- 
& \cellcolor{gray!20}-& \cellcolor{gray!20}-& \cellcolor{gray!20}- \\
Uniformerv2 \cite{uniformerv2}& IN-21K & 400 & 115                     
& -& -& - & - & - & - &-& -& 76.71 \\
\midrule

\multicolumn{13}{l}{\textbf{\textit{Image-to-Video PEFT Frameworks}}} \\

\cellcolor{gray!20}
& \cellcolor{gray!20}CLIP   & \cellcolor{gray!20}911 & \cellcolor{gray!20}& 
\cellcolor{gray!20}73.75 & \cellcolor{gray!20}95.71 & \cellcolor{gray!20}81.07 
& \cellcolor{gray!20}54.17 & \cellcolor{gray!20}77.84 & \cellcolor{gray!20}70.15 
& \cellcolor{gray!20}55.51 & \cellcolor{gray!20}88.36 & \cellcolor{gray!20}72.61 \\
\multirow{-2}{*}{\cellcolor{gray!20}ST-Adaptor \cite{St_adaptor}} 
& \cellcolor{gray!20}DINOv2 & \cellcolor{gray!20}1099 &  \multirow{-2}{*}{\cellcolor{gray!20}7.1}                                          
& \cellcolor{gray!20}81.50 & \cellcolor{gray!20}95.30 & \cellcolor{gray!20}85.45 
& \cellcolor{gray!20}63.48 & \cellcolor{gray!20}73.94 & \cellcolor{gray!20}70.54 
& \cellcolor{gray!20}66.43 & \cellcolor{gray!20}83.43 & \cellcolor{gray!20}75.19 \\

\multirow{2}{*}{M2-CLIP \cite{m2_clip}} 
& CLIP   & 935 & \multirow{2}{*}{14.8} 
& \underline{81.75} & \underline{95.07} & 85.85
& 64.11 & 76.62 & 71.86 
& 67.40 & \textbf{85.70} & 77.20 \\
& DINOv2 & 1128 &                      
& 78.39 & 84.64 & 80.47 
& \underline{68.49}& \underline{77.95} & \underline{74.88} 
&\underline{68.40}& 84.98& \underline{77.73} \\

\cellcolor{gray!20}
& \cellcolor{gray!20}CLIP   & \cellcolor{gray!20}937 & \cellcolor{gray!20}& 
\cellcolor{gray!20}60.53 & \cellcolor{gray!20}91.78 & \cellcolor{gray!20}70.95 
& \cellcolor{gray!20}50.11 & \cellcolor{gray!20}69.57 & \cellcolor{gray!20}63.25 
& \cellcolor{gray!20}65.45 & \cellcolor{gray!20}81.85 & \cellcolor{gray!20}73.98 \\
\multirow{-2}{*}{\cellcolor{gray!20}VitaCLIP \cite{VitaCLIP}} 
& \cellcolor{gray!20}DINOv2 & \cellcolor{gray!20}1130 &    \multirow{-2}{*}{\cellcolor{gray!20}28.6}                                       
& \cellcolor{gray!20}57.87 & \cellcolor{gray!20}53.21 & \cellcolor{gray!20}56.31 
& \cellcolor{gray!20}66.03 & \cellcolor{gray!20}74.97& \cellcolor{gray!20}72.48 
& \cellcolor{gray!20}65.45 & \cellcolor{gray!20}79.78 & \cellcolor{gray!20}72.91 \\

\midrule
\addlinespace[2pt]
\multicolumn{13}{l}{\textbf{\textit{Probing Frameworks (main baseline)}}} \\

\multirow{2}{*}{Linear Probing \cite{linear_prob_simclr} } 
& CLIP   & 269.6 & \multirow{2}{*}{\textbf{0.3}} 
& 19.46 & 35.35& 24.76  
& 25.29 & 36.39& 32.70 
& 28.77 & 66.27 & 48.28 \\
& DINOv2 & 351.5 &     
& 23.75 & 52.50 & 33.33 
& 32.45 & 46.15 & 41.71 
& 40.53 & 57.79 & 49.51 \\

\cellcolor{gray!20}
& \cellcolor{gray!20}CLIP   & \cellcolor{gray!20}273.6 & \cellcolor{gray!20}& 
\cellcolor{gray!20}30.71 & \cellcolor{gray!20}66.07 & \cellcolor{gray!20}42.50 
& \cellcolor{gray!20}60.77 & \cellcolor{gray!20}64.43& \cellcolor{gray!20}61.94 
& \cellcolor{gray!20}49.41 & \cellcolor{gray!20}78.69 & \cellcolor{gray!20}64.64 \\
\multirow{-2}{*}{\cellcolor{gray!20}Attn Probing \cite{set_transformer} } 
& \cellcolor{gray!20}DINOv2 & \cellcolor{gray!20}356.3 & \multirow{-2}{*}{\cellcolor{gray!20}7.3}                                          
& \cellcolor{gray!20}57.50 & \cellcolor{gray!20}72.85 & \cellcolor{gray!20}62.61 
& \cellcolor{gray!20}64.73 & \cellcolor{gray!20}70.64 & \cellcolor{gray!20}65.76 
& \cellcolor{gray!20}55.41 & \cellcolor{gray!20}80.16 & \cellcolor{gray!20}68.65 \\

\multirow{2}{*}{Self-Attn Probing \cite{clip4clip} } 
& CLIP   & 308.1 & \multirow{2}{*}{\underline{2.6}} 
& 45.35 & 77.85 & 56.19 
& 59.35 & 65.67 & 61.39 
& 51.22 & 79.48 & 65.93 \\
& DINOv2 & 413.5 &                      
& 74.28 & 80.00 & 76.19 
& 56.30 & 68.01 & 60.78 
& 59.57 & 80.85 & 71.16 \\

\cellcolor{gray!20}
& \cellcolor{gray!20}CLIP   & \cellcolor{gray!20}307.8 & \cellcolor{gray!20}
& \cellcolor{gray!20}56.25 & \cellcolor{gray!20}87.50& \cellcolor{gray!20}66.66 
& \cellcolor{gray!20}64.65& \cellcolor{gray!20}70.44& \cellcolor{gray!20}64.83 
& \cellcolor{gray!20}57.22 & \cellcolor{gray!20}80.37 & \cellcolor{gray!20}69.27 \\

\multirow{-2}{*}{\cellcolor{gray!20}\textbf{\textsc{Step} (Ours)}} 
& \cellcolor{gray!20}DINOv2 & \cellcolor{gray!20}413.1 & \multirow{-2}{*}{\cellcolor{gray!20}\underline{2.6}} 
& \cellcolor{gray!20}\textbf{82.14} & \cellcolor{gray!20}\textbf{96.78} & \cellcolor{gray!20}\textbf{87.02} 
& \cellcolor{gray!20}\textbf{69.46} & \cellcolor{gray!20}\textbf{80.59} & \cellcolor{gray!20}\textbf{76.80} 
& \cellcolor{gray!20}\textbf{69.98} & \cellcolor{gray!20}\underline{84.02} & \cellcolor{gray!20}\textbf{78.40} \\

\bottomrule
\end{tabular}
}}
\caption{Comparison of symmetric and non-symmetric action recognition accuracy, model efficiency, and trainable parameters. \textsc{Step} consistently outperforms fully fine-tuned, PEFT, and probing baselines.}
\vspace{-2em}
\label{table:symmetric_performance_with_params}
\end{table*}

\noindent\textbf{Learnable Global CLS Token.}
\textsc{Step} uses frame patch tokens and a \textbf{single learnable frame-global CLS} token, $e_{global}^{\text{CLS}}$ to maintain a coherent global representation. This differs from the standard way \cite{dosovitskiy2020vit,dinov2} of using a separate CLS token for each frame. We explicitly discard frame-specific CLS tokens and employ a single learnable global CLS token that attends to all the patch tokens across frames: 
\begin{equation}
e =  \left\{ e_{global}^{\text{CLS}}, e_1^{\text{patch}}, \ldots, e_T^{\text{patch}} \right\}
\label{eq:embedding_1}
\vspace{-0.3em}
\end{equation}
By attending to all patch tokens during self-attention, the global CLS token 
captures \textbf{global (sequence-level)} temporal dependencies. Meanwhile, \textbf{local (frame-level)} details are preserved by integrating frame-wise patch embeddings into the attention mechanism and classification layers, ensuring that \textsc{Step} maintains fine-grained spatial information while modeling the overall temporal structure. This design enhances temporal consistency by reducing redundancy across frames and consolidating the sequence into a single coherent representation, allowing the model to focus on key temporal transitions critical for action recognition.

\noindent\textbf{Injecting Temporal Embeddings in Self-attention Probing.}
To improve the temporal sensitivity of the existing probing mechanisms, we augment each frame representation $e_i$ with a learned frame-specific temporal embedding, denoted by  $t_i$. 
This results in a temporally enhanced frame embedding, $\tilde{e}_i$: $\tilde{e}_i = e_i + t_i$
However, since $\tilde{e}_i$ consists of multiple spatial patch tokens (Eq. \ref{eq:embedding_1}), temporal embedding $t_i$ is applied \textit{to each patch token} within the frame to ensure temporal information across all regions.
Thus, for each patch token $e_i^{\text{patch}}$ within frame $e_i$, the temporal embedding so added as  $\tilde{e}_i^{\text{patch}} = e_i^{\text{patch}} + t_i$, 
Consequently, self-attention probing is not permutation-invariant regarding the frame order, allowing the differentiation of actions with similar appearance but different temporal dynamics.

\noindent\textbf{Feature Aggregation and Classification.}
The embeddings  $\tilde{e}$  are passed through a Multihead Self-Attention (MHSA) layer followed by an average pooling operation: $p= \frac{1}{T} \sum_{t=1}^{T} \text{MHSA}(\tilde{e})$.
Unlike prior works~\cite{coca,video_glue,v-jepa,wang2024internvideo2scalingfoundationmodels}, we use a pure MHSA block without layer normalization, residual connections, or Feedforward (FF) layers, reducing the parameter count by approximately 3× while maintaining or slightly improving performance.
 While average pooling itself is permutation-invariant \cite{set_transformer}, the MHSA embeddings fed into it explicitly encode temporal order through frame-wise positional encodings and interactions with the global CLS token. Thus, the temporal order information is effectively preserved despite pooling. Finally, the pooled representation $p$ is passed to the linear classification layer.

\section{Experiments}
\label{sec:experiments}

\subsection{Evaluation setup}
\label{sec:setup}
We evaluate STEP using image-pretrained CLIP~\cite{CLIP} and DINOv2~\cite{dinov2} foundation models with ViT-B backbones across three diverse HRI-focused action recognition datasets.
HRI-30 is a recent benchmark specifically designed for human-robot interaction~\cite{hri_30}, while IKEA-ASM~\cite{IKEA_ASM} and Drive$\&$Act \cite{drive_and_act} focus on fine-grained hand-object assembly and in-car driver behavior, respectively. Both assembly and driving represent critical HRI domains where robots must interpret subtle, order-dependent human actions to provide safe and effective assistance. We adhere to standard evaluation protocols of prior work for fair comparison.


\noindent\textbf{Nearly Symmetric Actions Splits.} To study temporal order sensitivity, we also define \textit{nearly symmetric actions} as categories with visually similar frame appearance but nearly opposite temporal order.
We manually identify such categories in all three target datasets: HRI-30, IKEA-ASM, and Drive$\&$Act, each of which contains a substantial proportion of such actions.
Examples include  \textsl{pick up} vs. \textsl{lay down} or \textsl{open bottle} vs. \textsl{close bottle}. 
Our splits result in 20/14/20 (10/7/10 pairs) symmetric and 14/19/10 non-symmetric actions in  Drive$\&$Act, IKEA-ASM, and HRI-30 respectively. 

\subsection{Comparison to Probing and PEFT Baselines}
We first compare our model against two dominant adaptation paradigms for image-to-video transfer: probing and PEFT. Probing freezes the backbone with lightweight heads (linear~\cite{linear_prob_simclr}, attentive~\cite{coca,v-jepa}, self-attention~\cite{clip4clip}). These are efficient but largely permutation-invariant. PEFT methods (ST-Adaptor~\cite{St_adaptor}, Vita-CLIP~\cite{VitaCLIP}, M2-CLIP~\cite{m2_clip}) insert adapters or prompts for temporal modeling, but are heavier and remain underexplored on smaller domain-specific HRI datasets. Table \ref{table:symmetric_performance_with_params} reports accuracy for nearly symmetric, non-symmetric, and all categories, alongside FLOPs and trainable parameters to quantify both the performance and efficiency of the models. The results reveal three consistent findings.

\noindent\textbf{Large gains on symmetric actions.} \textsc{Step} consistently delivers the strongest results on \textit{nearly symmetric actions}. On IKEA-ASM, it improves symmetric action recognition by 4.7\% over attentive probing and still beats heavier PEFT methods (7–28M params). On HRI-30, \textsc{Step} boosts symmetric performance to 82.1\% (+7.8\% over probing), surpassing PEFT baselines even with far lower computational cost. Similarly, on Drive\&Act, \textsc{Step} raises symmetric accuracy over probing by +10.2\% to 69.98\%, while PEFT methods fall short despite significantly higher compute budgets. These results confirm that explicit temporal modeling is essential for symmetric actions, where probing and PEFT fall short.

\noindent\textbf{Strong performance on non-symmetric and overall accuracy.} Importantly, \textsc{Step} does not trade off symmetric accuracy for non-symmetric classes. On HRI-30, it achieves 96.8\% non-symmetric and 87.02\% overall accuracy, the best across all methods. On IKEA-ASM, it attains 80.6\% non-symmetric and 76.8\% overall, outperforming probing baselines, surpassing PEFT methods (e.g., M2-CLIP at 74.9\%). On Drive\&Act, PEFT closes the gap on non-symmetric actions (85.7\% for M2-CLIP), yet \textsc{Step} remains competitive with 84.0\% non-symmetric and the strongest overall balance (78.4\%). In summary, \textsc{Step} is effective for both symmetric and non-symmetric actions, with the strongest gains observed for the symmetric ones, which are especially prominent in HRI tasks comprising fine-grained object manipulation. 

\noindent\textbf{Efficiency and robustness across backbones.} 
A key strength of \textsc{Step} is that it enables efficient VFM adaptation while delivering state-of-the-art performance.
With only 2.6M trainable parameters and 410 GFLOPs, \textsc{Step} is an order of magnitude smaller than typical PEFT frameworks (7-28M params and ~900–1100 GFLOPs per pass). 
Although CNN-based task-specific models remain more lightweight, VFMs offer superior recognition quality and multi-task capabilities through shared frozen backbones (Table \ref{tab:multitask}).
Compared to probing, \textsc{Step} adds only ~1-2\% compute but delivers dramatic accuracy gains: +5 to +20  points on symmetric and overall recognition. These improvements are consistent across both CLIP and DINOv2 backbones. 

\noindent\textbf{Comparison to fully fine-tuned.} 
Previous approaches rely on fully fine-tuned heavy video backbones such as VideoSWIN \cite{liu2022video_swin}, SlowOnly, and UniformerV2 \cite{uniformerv2}, which represent the current SOTA on these benchmarks. 
These models achieve strong results (e.g., 86.6\% overall on HRI-30 and 76.7\% on Drive\&Act) but at the cost of more trainable parameters, and they are highly task-specific.  In contrast, \textsc{Step} surpasses these models with minimal fine-tuning, showing that probing frozen VFMs can rival or exceed full fine-tuning while remaining more generalizable across tasks and domains.
\subsection{Temporal Order Sensitivity Analysis}
\label{sec:Temp_sensitivity}
\noindent\textbf{Impact of Test-time Frame Order Corruptions.}
Temporal order is generally an important element defining human activities, and a reliable action recognition model should show reduced performance if the event sequence is altered at test time. To evaluate this, we test probing and PEFT methods under correct and reverse frame order (Table \ref{tab:temporal_sensitivity}). 
 




\begin{table}[ht]
\centering
\resizebox{\columnwidth}{!}{%
{\fontsize{7}{9}\selectfont
\setlength{\tabcolsep}{4pt}
\begin{tabular}{l | cc | cc}
\toprule
\textbf{Dataset} & \multicolumn{2}{c|}{\cellcolor{gray!25}\textbf{Probing}} & \multicolumn{2}{c}{\cellcolor{gray!25}\textbf{PEFT}} \\
& \textbf{Attn.} & \textbf{\textsc{Step}(Ours)} & \textbf{ST-Adaptor} & \textbf{M2-CLIP}\\
\midrule
HRI-30            & \multirow{2}{*}{62.61}& 87.02& 85.45& 80.47\\
HRI-30\_Reversed  & &42.26 (\textcolor{red}{$\downarrow$} 44.76) & 39.76 (\textcolor{red}{$\downarrow$} 45.69) & 37.38 \textcolor{red}(\textcolor{red}{$\downarrow$} 43.09) \\
\midrule
IKEA-ASM          & \multirow{2}{*}{65.76}& 76.28 & 70.54& 74.88\\
IKEA-ASM\_Reversed & & 55.19(\textcolor{red}{$\downarrow$} 21.1) & 67.28 (\textcolor{red}{$\downarrow$} 3.3) & 68.91(\textcolor{red}{$\downarrow$} 5.9) \\

\midrule
Drive\&Act        & \multirow{2}{*}{68.65}& 78.40 & 75.19& 77.73\\
Drive\&Act\_Reversed & & 59.83 (\textcolor{red}{$\downarrow$} 18.6)& 61.16 (\textcolor{red}{$\downarrow$} 14.03) &59.44 (\textcolor{red}{$\downarrow$} 18.2) \\
\bottomrule
\end{tabular}
}}
\caption{Comparison of probing (Attentive probing vs. \textsc{STEP}) and PEFT methods with and without reversed frames at test time across HRI-30, Drive\&Act, and IKEA-ASM.}
\label{tab:temporal_sensitivity}
\vspace{-0.5em}
\end{table}

Conventional probing methods remain unaffected in the reverse configuration, confirming their inherent permutation invariance and inability to model temporal order. 
In contrast, both PEFT methods and \textsc{Step}  show clear sensitivity to temporal corruption, with large drops in the reversed setting (e.g., –44.8\% on HRI-30 for \textsc{Step}). This demonstrates that \textsc{Step}, despite operating only at the probing stage with 2.6M parameters, encodes order dependencies comparably to or better than heavy PEFT frameworks.

\noindent\textbf{Class-wise symmetric action breakdown}
We further analyze the performance of \textsc{Step} on individual \textit{nearly symmetric actions} of the Drive$\&$Act dataset. Table \ref{table:issues_symmetric} presents their overall accuracy with the attentive probing baseline, the activity class most frequently confused with, and the corresponding confusion rate. Interestingly, we see that the model tends to reliably recognize \textit{one} action in a symmetric pair, while the other is often mapped to its nearly symmetric counterpart. For example, \textsl{closing bottle} has a correct recognition rate of only $17\%$ cases and is confused with \textsl{opening bottle} in $60\%$ of cases. 
\textsl{Opening bottle} on the other hand is correctly recognized $68\%$ of times.
With our approach, identifying \textsl{closing bottle} works $29\%$ better and the confusion with \textsl{opening bottle} falls by $20\%$.
This pattern is repeated for most of the pairs.
\newcommand{\good}[1]{\cellcolor{green!70}#1}
\newcommand{\average}[1]{\cellcolor{green!25}#1}
\newcommand{\bad}[1]{\cellcolor{red!60}#1}
\newcommand{\better}[1]{\cellcolor{red!20}#1}

\begin{table}[ht!]
\centering

\resizebox{\columnwidth}{!}{%
\begin{tabular}{lcllcl}

\toprule
\textbf{True Activity } & \multicolumn{2}{c}{\textbf{Top-1 Acc} }&\textbf{Most Common } & \multicolumn{2}{c}{\textbf{Confusion} }\\
\textbf{Class} & \textbf{Attn.}&  \textbf{$\Delta$} ({$\uparrow$})&\textbf{Confusion Class} &   \textbf{Attn.}&\textbf{$\Delta$} ({$\downarrow$})\\
\midrule
closing\_bottle & 0.17&  \good{0.29}&opening\_bottle & 0.60&\good{-0.20}\\ 
closing\_door\_inside & 0.94&  -0.12&opening\_door\_inside & 0.06&\better{0.12}\\ 
closing\_door\_outside & 0.73&  \good{0.27}&opening\_door\_outside & 0.27&\good{-0.27}\\ 
closing\_laptop & 0.22&  \average{0.06}&Working on laptop & 0.28&\average{-0.06}\\ 
entering\_car & 0.94&  \average{0.06}&exiting\_car & 0.06&\average{-0.06}\\ 
exiting\_car & 1&  0.0&- & 0.0&0.0\\ 
fastening\_seat\_belt & 0.89&  \average{0.05}&unfastening\_seat\_belt & 0.08&\average{-0.06}\\ 
fetching\_an\_object & 0.51&  \good{0.32}&placing\_an\_object & 0.37&\good{-0.29}\\ 
opening\_bottle & 0.68&  \average{0.08}&eating & 0.11&\better{0.05}\\ 
opening\_door\_inside & 0.19&  \good{0.56}&closing\_door\_inside & 0.56&\good{-0.50}\\ 
opening\_door\_outside & 0.71&  \good{0.29}&closing\_door\_outside & 0.29&\good{-0.29}\\ 
opening\_laptop & 0.24&  -0.12& working on laptop & 0.47&\good{-0.24}\\ 
placing\_an\_object & 0.62&  \average{0.18}&fetching\_an\_object & 0.22&\average{-0.12}\\ 
putting\_laptop\_into\_bag & 0.14&  -0.14&placing\_an\_object & 0.29&\bad{0.43}\\ 
putting\_on\_jacket & 0.33&  \average{0.05}&taking\_off\_jacket & 0.24&\better{0.10}\\ 
putting\_on\_sunglasses & 0.84&  -0.28&talking\_on\_phone & 0.04&\better{0.20}\\ 
taking\_laptop\_from\_bag & 0.43&  0.0&placing\_an\_object & 0.43&\average{-0.10}\\ 
taking\_off\_jacket & 0.67&  -0.27&placing\_an\_object & 0.20&\bad{0.27}\\ 
taking\_off\_sunglasses & 0.22&  \average{0.13}&putting\_on\_sunglasses & 0.22&\average{-0.04}\\ 
unfastening\_seat\_belt & 0.5&  \good{0.25}&fastening\_seat\_belt & 0.25&\average{-0.11}\\
\bottomrule
\end{tabular}%
}
\caption{Analysis of the \textit{nearly symmetric actions} of the Drive$\&$Act dataset, including accuracy and most common confusions. $\Delta$ showcases comparison to  \textsc{Step}.
}
\vspace{-1em}
\label{table:issues_symmetric}
\end{table}

\begin{figure}[ht!]
    \centering
\includegraphics[width=0.95\columnwidth]{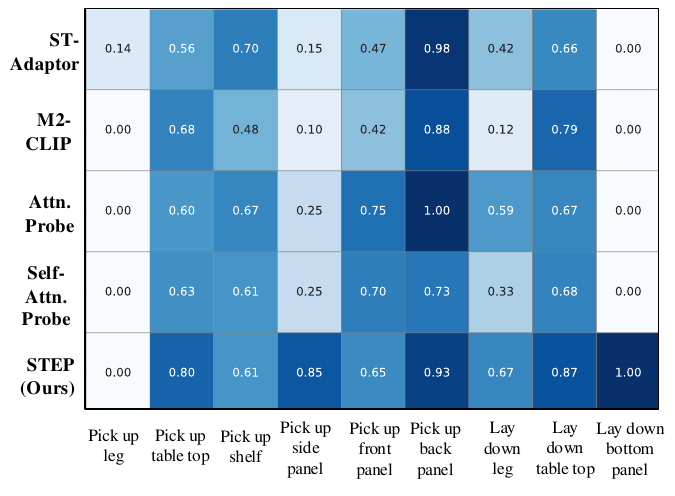}
\vspace{-1em}
\caption{Class-wise accuracy on nearly symmetric actions in IKEA-ASM. STEP outperforms PEFT  and probing baselines.
}
\vspace{-1em}
\label{fig:ikea_symm}
\end{figure}

We also perform a similar analysis on the IKEA-ASM dataset, especially with the PEFT methods.  As shown in Figure \ref{fig:ikea_symm}, \textsc{Step} consistently outperforms PEFT methods across nearly all actions, with only a few cases where other methods perform comparably. Despite these exceptions, \textsc{Step} maintains better performance overall, particularly in distinguishing nearly symmetric actions.

\subsection{Multi-Task Performance}

 \begin{figure}[ht!]
    \centering
\includegraphics[width=0.98\columnwidth]{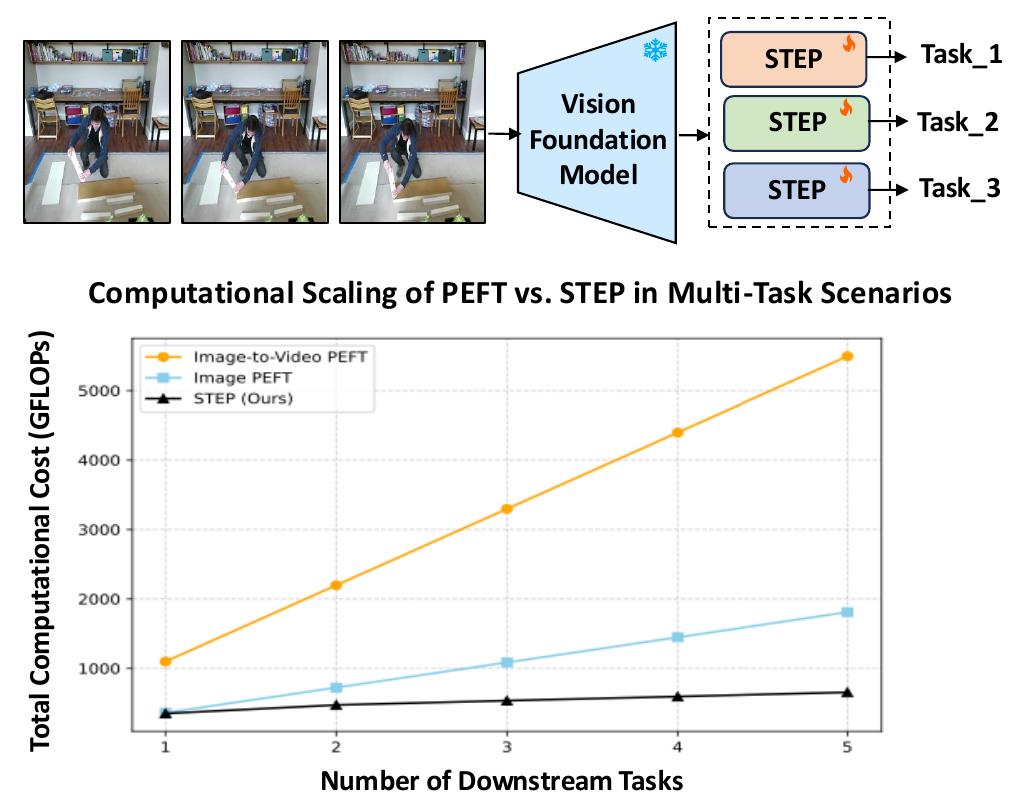}
\vspace{-0.5em}
\caption{Computational scaling of PEFT vs. \textsc{Step} during inference in multi-task scenarios. PEFT cost grows linearly with tasks, while \textsc{Step} remains constant. }
\vspace{-1.4em}
\label{fig:multitask_scaling}
\end{figure}

Robotic systems must often perform several perception tasks at once, such as fine-grained activity recognition (FAR), atomic action recognition (AAR), and identifying the object under interaction (OUI), to provide intention-aware assistance. Existing PEFT methods (e.g., ST-Adaptor, VPT) train adapters or prompts separately for each task, forcing multiple backbone passes and scaling inference cost linearly with task count (Fig. \ref{fig:multitask_scaling}), even when trained under identical training budgets. Even when forced into a single-pass setup by reusing a single PEFT backbone across tasks, performance drops notably because the backbone overfits to the optimization task, hurting generalization on others (Table \ref{tab:multitask}).
\textsc{Step} avoids this by sharing a frozen VFM backbone across tasks and attaching lightweight probes, enabling all objectives in a single pass and avoiding redundant backbone forward passes required by task-specific PEFT adapters. On IKEA-ASM, \textsc{Step} reduces inference cost (GFLOPs) by up to 6× while improving accuracy across FAR, AAR, and OUI tasks. Moreover, adding new tasks only requires probes, not backbone re-optimization, making \textsc{Step} highly scalable for real-time HRI.

\begin{table*}[ht!]
\centering
{\fontsize{8}{10}\selectfont
\setlength{\tabcolsep}{4pt}

\begin{minipage}{0.63\linewidth} 
\centering
\resizebox{\linewidth}{!}{%
\begin{tabular}{l c |ccc  |ccc  |ccc }
\toprule
\textbf{Method} & \textbf{Classifier} &
 \multicolumn{3}{c}{\cellcolor{red!25}\textbf{HRI-30}} & 
 \multicolumn{3}{c}{\cellcolor{blue!25}\textbf{IKEA-ASM}} & 
 \multicolumn{3}{c}{\cellcolor{green!25}\textbf{Drive\&Act}}\\
& & \textbf{Sym} & \textbf{N-Sym} & \textbf{Ovr.} & 
\textbf{Sym} & \textbf{N-Sym} & \textbf{Ovr.} & 
\textbf{Sym} & \textbf{N-Sym} & \textbf{Ovr.} \\
\midrule
\multirow{3}{*}{VPT\cite{vpt}} & Linear  & 51.07& 85.53& 62.50 & 60.62 &73.82 & 69.54 & 56.04 & 79.98 & 68.49 \\
&  \textsc{Step}    & 74.11 & \textbf{96.07}& 81.43 & \textbf{70.88}& \textbf{78.30}&\textbf{75.89} & \textbf{72.72} & 82.64 & 77.88 \\
&  $\Delta (\uparrow)$  & \cellcolor{green!45}+23.07 & \cellcolor{green!45}+10.54 & \cellcolor{green!45}+18.93
& \cellcolor{green!45}+10.26 & \cellcolor{green!15}+4.48 & \cellcolor{green!15}+6.35
& \cellcolor{green!45}+16.68 & \cellcolor{green!15}+2.66 & \cellcolor{green!15}+9.39 \\
\midrule
\multirow{3}{*}{AdaptFormer\cite{adaptformer}} & Linear  & 64.28 & 76.79 & 68.45 & 65.16 & 73.13& 70.54 & 60.43 & 83.99 & 73.73 \\
&  \textsc{Step}    & \textbf{75.53} &93.58  &\textbf{81.55} & 65.66 & 76.11& \textbf{72.56} & 70.59& \textbf{85.99}& \textbf{78.61} \\
&  $\Delta (\uparrow)$  & \cellcolor{green!45}+11.25 & \cellcolor{green!45}+16.78 & \cellcolor{green!45}+13.09
& \cellcolor{green!15}+0.50 & \cellcolor{green!15}+2.98 & \cellcolor{green!15}+2.02 
& \cellcolor{green!45}+10.16 & \cellcolor{green!15}+2.00 & \cellcolor{green!15}+4.88 \\
\bottomrule
\end{tabular}}
\caption{Comparison of \textsc{Step} with image PEFTs on HRI-30, IKEA-ASM, and Drive\&Act. \textsc{Step} consistently improves symmetric and overall accuracy.}
\label{tab:step_image_peft}
\end{minipage}
\hfill
\begin{minipage}{0.35\linewidth} 
\centering
\resizebox{\linewidth}{!}{%
\renewcommand{\arraystretch}{1.2}
\begin{tabular}{l c c | ccc}
\toprule
\textbf{Method} & \textbf{\# Passes} & \textbf{GFLOPs} &
\textbf{FAR} & 
\textbf{AAR} & 
\textbf{OUI} \\
& & & 
\textbf{Acc.} & 
\textbf{Acc.} & 
\textbf{Acc.} \\
\midrule
\cellcolor{gray!20}& \cellcolor{gray!20}1 & \cellcolor{gray!20}1099  & \cellcolor{gray!20}52.63 & \cellcolor{gray!20}47.83 & \cellcolor{gray!20}74.26 \\
\cellcolor{gray!20}\multirow{-2}{*}{ST-Adaptor \cite{St_adaptor}}  
&\cellcolor{gray!20}3 & \cellcolor{gray!20}3297  & \cellcolor{gray!20}70.54 & \cellcolor{gray!20}72.95 & \cellcolor{gray!20}74.26 \\
\multirow{2}{*}{VPT \cite{vpt}}  
& 1 & \textbf{362.4} & 56.05 & 54.03 & 66.28 \\
& 3 & 1087  & 69.54 & 65.82 & 66.28 \\
\rowcolor{gray!10}
\textbf{STEP (Ours)}           
& \textbf{1} & \textbf{536.5} & \textbf{76.80} & \textbf{75.28} & \textbf{76.51} \\
\bottomrule
\end{tabular}}
\caption{Multi-task: \textsc{Step} achieves higher accuracy with less GFLOPs.
}
\label{tab:multitask}
\end{minipage}
}
\vspace{-1em}
\end{table*}

\subsection{\textsc{Step} as Temporal Extension of Image PEFT}

While image-to-video PEFTs \cite{St_adaptor, m2_clip} perform strong spatio-temporal modeling, they are computationally heavy, requiring 7–28M parameters and high GFLOPs. In contrast, image-based PEFTs like Visual Prompt Tuning \cite{vpt} (VPT) and AdaptFormer \cite{adaptformer} are lightweight but focus mainly on spatial features, ignoring temporal reasoning. As shown in Table~\ref{tab:step_image_peft}, adding \textsc{Step} consistently boosts their performance across all datasets, with large gains on symmetric actions (+23.1\% for VPT on HRI-30, +10.1\% for AdaptFormer on Drive\&Act) and up to +18.9\% overall accuracy. Notably, \textsc{Step}+image PEFT surpasses heavy video PEFTs on IKEA-ASM and Drive\&Act, while approaching standalone \textsc{Step}. Still, \textsc{Step} alone delivers the best overall performance, establishing state-of-the-art results and highlighting its versatility as a temporal plug-in for lightweight Image-based PEFTs.

\begin{table}[t]
\centering
\begin{minipage}{0.54\linewidth} 
    \centering
    \resizebox{\columnwidth}{!}{
    \begin{tabular}{l c c }
    \toprule
    \textbf{Method} & \textbf{D\&A} & \textbf{IKEA} \\
    \midrule
    FF + LN + Skip (Block) & 77.27 & 73.72 \\
    LN + Skip (No FF) & 76.55 & 74.88 \\
    Ours (only Attn. layer) & \textbf{78.40} & \textbf{76.28} \\
    \bottomrule
    \end{tabular}
    }
    \caption{Comparison of attention block variants.}
    \label{table:attn_layer_vs_block}
\end{minipage}
\hfill
\begin{minipage}{0.44\linewidth} 
    \centering
    \resizebox{\columnwidth}{!}{
    \begin{tabular}{l c c}
    \toprule
    \textbf{Method} & \textbf{D\&A} & \textbf{IKEA} \\
    \midrule
    Only Global CLS & 70.19 & 54.11 \\
    Only Patch Tokens & 77.29 & 73.49 \\
    Combined (\textsc{Step}) & \textbf{78.40} & \textbf{76.28} \\
    \bottomrule
    \end{tabular}
    }
    \caption{Ablation of CLS vs. Patch Tokens.}
    \label{table:cls_patch}

\end{minipage}
\vspace{-1em}
\end{table}

\subsection{Qualitative Comparisons}
\begin{figure}[ht!]
    \centering
\includegraphics[width=1\columnwidth]{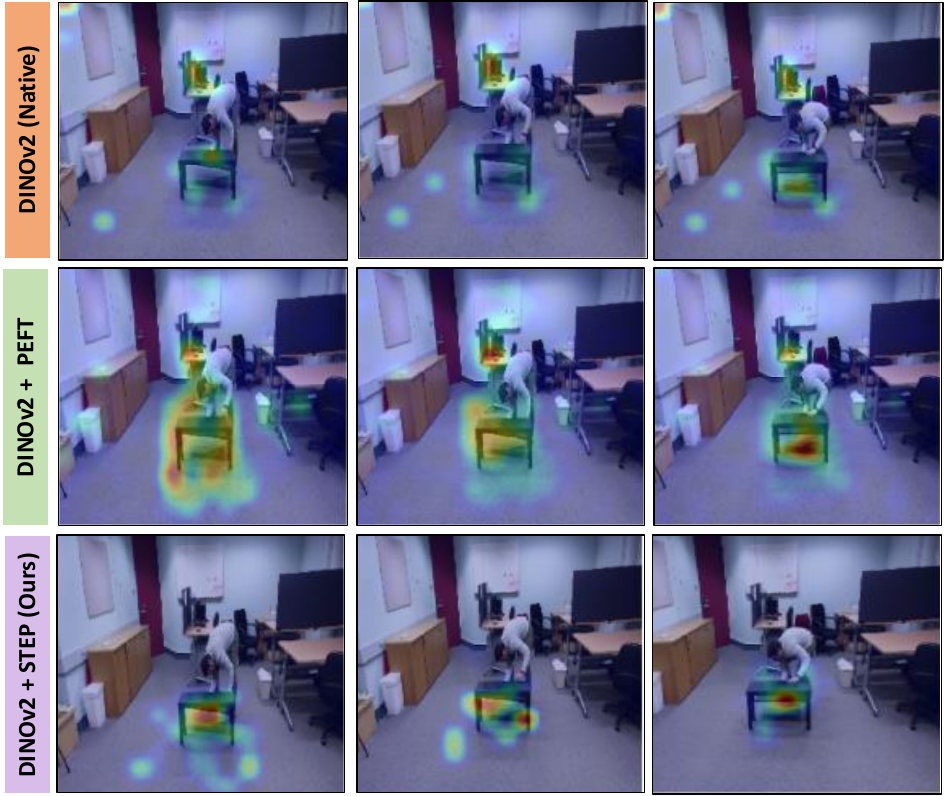}
\caption{Attention maps for action recognition. DINOv2 attends to background, PEFT spreads attention diffusely, while our DINOv2 + \textsc{Step} focuses on the human–object interaction, aiding discrimination of symmetric actions. }
\label{fig:qualitative}
\vspace{-0.5em}
\end{figure}

To better understand how \textsc{Step} improves recognition, we visualize attention maps (Fig.~\ref{fig:qualitative}). Frozen DINOv2 largely attends to the background, overlooking fine-grained cues. PEFT shifts focus toward the human and furniture but still diffuses attention to irrelevant background objects. In contrast, DINOv2 + \textsc{Step} sharply concentrates on the human–object interaction (e.g., the hand and manipulated object), indicating that our lightweight probe explicitly drives attention toward sequence-relevant regions.

 \begin{figure}[ht!]
    \centering
\includegraphics[width=1\columnwidth]{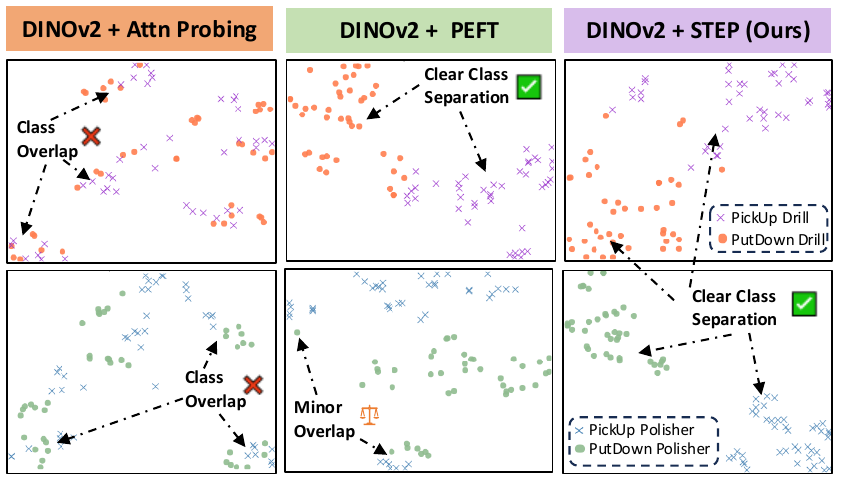}
\caption{t-SNE visualizations of HRI-30 dataset. Unlike probing (overlap) and PEFT (minor overlap), DINOv2 + \textsc{Step} cleanly separates\textit{ nearly symmetric actions}, enabling safer and more reliable action understanding in HRI.  }
\label{fig:t_sne}
\vspace{-1em}
\end{figure}

We further analyze class separability through t-SNE embeddings of \textit{nearly symmetric action} pairs such as \emph{pick up drill} vs.  \emph{put down drill} (Figure \ref{fig:t_sne}). 
The results reinforce our earlier observations: attentive probing produces almost complete overlap between classes, confirming its permutation-invariant nature and inability to capture temporal order. PEFT methods achieve improved separation, demonstrating they perform temporal modeling, yet residual overlaps remain, reflecting their reduced effectiveness in symmetric actions. In contrast, \textsc{Step} produces sharp, well-isolated clusters with minimal confusion, showing our lightweight probe effectively encodes sequence direction and disambiguates visually similar but temporally opposite actions.

\subsection{Ablation Studies}
\noindent\textbf{Impact of Simplified Attention block.}
We analyze the effect of simplifying the probing attention layer by progressively removing feedforward layers, normalization, and residual connections. These components mainly stabilize deep transformers but add unnecessary overhead in our single-layer setup. Table~\ref{table:attn_layer_vs_block} shows that our simplified attention block, retaining only multi-head attention, improves performance while reducing complexity, achieving 78.40\% on Drive\&Act (+1.13\%), and 76.28\% on IKEA-ASM (+1.93\%). This demonstrates that a minimal design not only reduces complexity but also enhances accuracy in probing.

\noindent\textbf{Impact of Global CLS Token and Frame-wise Temporal Positional Encoding.}
We analyze the contributions of \textsc{Step}'s components by evaluating the Global CLS Token and Frame-wise Temporal Positional Encoding (PE). Table~\ref{table:ablation_study} shows that adding the Global CLS token consistently improves accuracy, with the largest +8\% gain on IKEA-ASM, where capturing global context is critical. The Frame-wise Temporal PE further boosts accuracy on both Drive\&Act (+5.91\%) and IKEA-ASM (+10.78\%), reinforcing its importance for nearly symmetric actions.
In Table~\ref{table:cls_patch}, relying on CLS alone underperforms on fine-grained datasets where local detail is crucial, whereas patch embeddings complement CLS by capturing low-level cues. Combining both, as in \textsc{Step}, yields the strongest results across all benchmarks.

\begin{table}[t]
\centering
\begin{minipage}{0.48\linewidth} 
    \centering
    \resizebox{\columnwidth}{!}{
    \begin{tabular}{l c c }
    \toprule
    \textbf{Method} & \textbf{D\&A} & \textbf{IKEA} \\
    \midrule
    Fixed PE & 76.76 & 69.53 \\
    Hybrid PE & 77.78 & 74.88 \\
    Learnable PE (Ours) & \textbf{78.40} & \textbf{76.28} \\
    \bottomrule
    \end{tabular}
    }
    \caption{Ablation of different PE types.}
    \label{tab:PE}
\end{minipage}
\hfill
\begin{minipage}{0.48\linewidth} 
    \centering
    \resizebox{\columnwidth}{!}{
    \begin{tabular}{l c c}
    \toprule
    \textbf{Method} & \textbf{D\&A} & \textbf{IKEA} \\
    \midrule
    Token-wise PE & 73.21 & 68.37 \\
    Frame-wise PE (\textsc{Step}) & \textbf{78.40} & \textbf{76.28} \\
    \bottomrule
    \end{tabular}
    }
    \caption{Ablation of token vs. frame-wise PE.}
    \label{table:token_pe}
\end{minipage}
\vspace{-1.4em}
\end{table}
\noindent\textbf{Impact of Different Positional Encoding Schemes.}
\begin{table}[t!]
\centering
\begin{minipage}{\columnwidth}
\resizebox{\columnwidth}{!}{
\begin{tabular}{l cc c }
 
\toprule

\textbf{Method} & \textbf{HRI-30}& \textbf{Drive\&Act} & \textbf{IKEA-ASM} \\

\midrule
 \multicolumn{4}{l}{\textbf{Global CLS Token}}\\
 Self-Attn. Probing& 76.19& 71.16& 60.78\\
 Self-Attn. Probing w Global CLS& 78.88& 72.00&68.76\\
 \midrule
 \multicolumn{4}{l}{\textbf{Frame-wise Temporal PE}}\\
 Self-Attn. Probing w Temporal PE& 84.74& 77.07&71.56\\
 \textsc{Step}& \textbf{87.02}& \textbf{78.40}&\textbf{76.28}\\
 \bottomrule
\end{tabular}
}
\end{minipage}
\caption{Ablation of individual components and comparison to self-attention probing without our modifications. 
}
\vspace{-2em}
\label{table:ablation_study}
\end{table}
Table~\ref{tab:PE} evaluates different positional encodings for DINOv2. Learnable PE, used in \textsc{Step}, consistently outperforms Fixed and Hybrid variants, with the largest gains on IKEA-ASM. We further compare our frame-wise PE with conventional token-wise PE (Table~\ref{table:token_pe}) and find that, despite using fewer parameters, it achieves higher accuracy, highlighting its effectiveness in capturing temporal order.

\section{Conclusion and Limitations}
\label{sec:conclusion}

We studied parameter-efficient image-to-video probing, focusing on \textit{nearly symmetric actions} -- visually similar actions unfolding in reverse order. Existing approaches struggle with such actions as they ignore frame order due to permutation-invariant attention. To address this, we proposed \textsc{Step}, introducing simple yet effective modifications to self-attention probing, improving temporal sensitivity. 
\textsc{Step} outperforms probing, PEFT and fully fine-tuned methods across HRI-30, IKEA-ASM, and Drive\&Act, achieving state-of-the-art accuracy with only a fraction of the trainable parameters. It also enables multi-task inference in a single backbone pass, reducing computation up to 6$\times$ compared to PEFT. Limitations remain: inference still requires a full pass through large transformer backbones; and while frozen probing excels in today’s small-scale HRI benchmarks, its advantage may shrink if large symmetric-action datasets become available, where full fine-tuning and PEFT could dominate. Still, in realistic HRI scenarios, where data and compute are limited, \textsc{Step} offers a practical balance of efficiency and accuracy.

\section*{Acknowledgments}
The research published in this article is supported by the Deutsche Forschungsgemeinschaft
(DFG) under Germany’s Excellence Strategy – EXC 2120/1 –390831618. 
The authors also thank the International Max Planck Research School for Intelligent Systems (IMPRS-IS) for supporting Thinesh Thiyakesan Ponbagavathi.
The authors also gratefully acknowledge the computing time provided on the high-performance computer HoreKa by the National High-Performance Computing Center at KIT. 
HoreKa is partly funded by the German Research Foundation (DFG).

{\small
\bibliographystyle{IEEEtran}
\bibliography{IEEEabrv,IEEEexample}


\end{document}